\documentclass[letterpaper,10pt, conference]{ieeeconf}
\IEEEoverridecommandlockouts      
\usepackage{graphics} 
\usepackage{graphicx} %
\usepackage{gensymb}

\usepackage{tikz}
\usepackage{verbatim}
\usetikzlibrary{calc}

\usepackage{amsmath} 
\usepackage{amssymb}  
\usepackage{mathtools}
\usepackage{algorithmic}

\usepackage{dirtytalk}

\usepackage[utf8]{inputenc}
\usepackage{cite}
\usepackage{textcomp}
\usepackage{xcolor}
\usepackage{subcaption}
\usepackage{soul}
\usepackage{tikz,xcolor}
\usepackage[font=footnotesize]{caption}

\usepackage[colorlinks,citecolor=red,urlcolor=black,bookmarks=false,hypertexnames=true]{hyperref}

\newcommand*{\Ye}{\textcolor{blue}}

\def\BibTeX{{\rm B\kern-.05em{\sc i\kern-.025em b}\kern-.08em
    T\kern-.1667em\lower.7ex\hbox{E}\kern-.125emX}}

\title{\LARGE \bf Hierarchical Experience-informed Navigation for Multi-modal Quadrupedal Rebar Grid Traversal}

\author{{Max Asselmeier$^{1}$, Jane Ivanova$^{2}$, Ziyi Zhou$^{3}$, Patricio A. Vela$^{3}$ and Ye Zhao$^{1}$ }
\thanks{$^{1}$M. Asselmeier and Y. Zhao are with the School of Mechanical Engineering, Georgia Institute of Technology, Atlanta, GA, 30308, USA. {\tt\small \{mass, yzhao301\}@gatech.edu}}%
\thanks{$^{2}$ J. Ivanova is with SkyMul Inc, Atlanta, GA, 30339, USA. {\tt\small jane.ivanova.work@gmail.com }}
\thanks{$^{3}$ Z. Zhou and P.A. Vela are with the School of Electrical and Computer Engineering, Georgia Institute of Technology, Atlanta, GA, 30308, USA. {\tt\small \{zhouziyi, pvela\}@gatech.edu}}
\thanks{The work of Max Asselmeier is supported by the National Science Foundation Graduate Research Fellowship under Grant No. DGE-2039655. Any opinion, findings, and conclusions or recommendations expressed in this material are those of the authors(s) and do not necessarily reflect the views of the National Science Foundation. This work is also supported through a sponsored research grant from SkyMul Inc. and a Georgia Tech IRIM/IPaT Aware Home Seed Grant. }}%

\begin{document}

\maketitle
\thispagestyle{empty}
\pagestyle{empty}


\begin{abstract}
This study focuses on a layered, experience-based, multi-modal contact planning framework for agile quadrupedal locomotion over a constrained rebar environment. To this end, our hierarchical planner incorporates locomotion-specific modules into the high-level contact sequence planner and performs kinodynamically-aware trajectory optimization as the low-level motion planner. Through quantitative analysis of the experience accumulation process and experimental validation of the kinodynamic feasibility of the generated locomotion trajectories, we demonstrate that the planning heuristic of experience offers an effective way of providing candidate footholds for a legged contact planner. Additionally, we introduce a guiding torso path heuristic at the global planning level to enhance the navigation success rate in the presence of environmental obstacles. Our results indicate that the torso-path guided experience accumulation requires significantly fewer offline trials to successfully reach the goal compared to regular experience accumulation. Finally, our planning framework is validated in both dynamics simulations and real hardware implementations on a quadrupedal robot provided by Skymul Inc.
\end{abstract}

\section{Introduction} \label{sec:Introduction}

The task of legged locomotion elicits a hybridized planning space that combines discrete elements like robot end effectors and environmental artifacts that can support footsteps along with continuous footstep positions along such artifacts. Methods that opt to perform simultaneous contact and footstep position planning often struggle to do so in a computationally tractable manner because of this hybrid planning space. The need for kinodynamically feasible solutions to this planning problem further compounds the computational efforts required.

An alternative way to resolve contact planning is through a hierarchical approach in which a discrete contact sequence is generated at the higher level and then continuous whole-body trajectories that abide by the generated contact sequence are synthesized at the lower level. The separation of the planning space into discrete and continuous components computationally simplifies the overall planning problem.  

This planning decomposition reflects those widely explored in the areas of task and motion planning (TAMP) \cite{toussaint_differentiable_2018, hartmann_long-horizon_2023,kim_reachability_2023} and multi-modal motion planning (MMMP) \cite{hauser_non-gaited_2005, kingston_scaling_2023, bretl_motion_2006}. These areas have proposed numerous effective planning heuristics that allow the discrete and continuous planning layers to inform each other and coordinate useful planning attempts. However, traditional MMMP has seen limited use in dynamic locomotion due the combinatorial nature of contacts which makes the problem difficult to scale up and the inherently dynamic process of legged locomotion which imposes complex constraints on motion planning.

\begin{figure}[t!]
    \centering
    \includegraphics[width=0.48\textwidth]{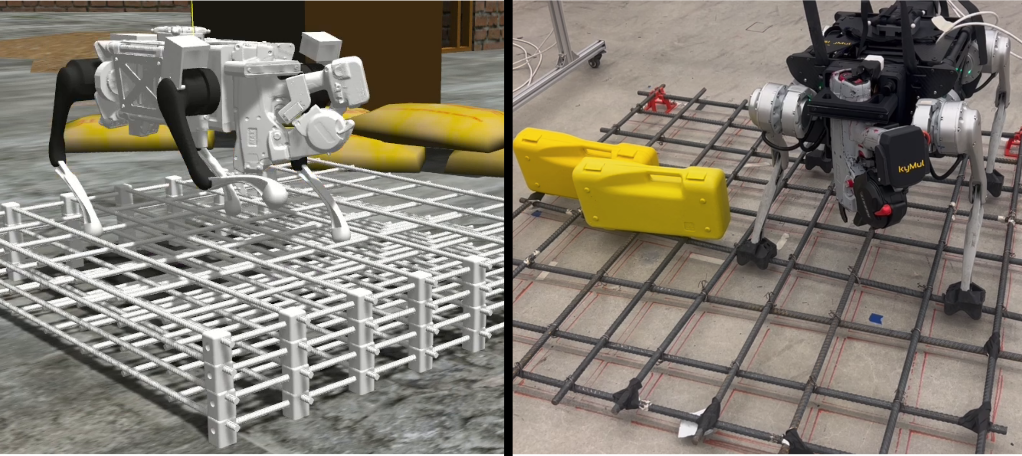}
    \caption{Illustration of a quadrupedal robot performing rebar grid traversal in a simulated construction environment \cite{clearpath_robotics_clearpath_nodate} and an indoor test-bed. }
    \label{fig:go1_rebar}
    \vspace{-1.0cm}
\end{figure}

In this work, we draw inspiration from the Augmented Leafs with Experience on Foliations (ALEF) framework \cite{kingston_scaling_2023} for multi-modal planning. In particular, we design a novel hierarchical planning framework to generate kinodynamically-feasible quadrupedal locomotion plans, taking into account the robot's centroidal dynamics and kinematic reachability. This study marks the first effort that leverages model-based trajectory optimization (TO) in the design of the experience heuristic for quadrupedal locomotion.
Our main contributions are summarized as below:
\begin{enumerate}
    \item Adapting the concept of a mode transition graph to quadrupedal contact planning along with a carefully designed experience heuristic to weight the mode transition graph and guide contact sequence planning;
    \item Integrating mode transition graph search with lower-level TO to naturally embed footstep planners and tightly integrate the kinodynamically-aware optimal cost into the experience heuristic;
    \item Integrating this multi-modal contact planner into a navigation framework to exploit a guiding torso path;
    \item Experimental validation of the proposed framework for rebar grid traversal through quantitative analysis of both simulations and hardware implementation.
\end{enumerate}

\section{Related Work}
\label{sec:relatedwork}

\subsection{Contact Planning}
Within the quadrupedal contact planning domain, there exists a trade-off between solution quality and computational cost. In the simplest case, fixing contact schedules and generating nominal footstep positions through Raibert heuristics \cite{raibert_legged_1986} allows for online planning at high frequencies \cite{bledt_implementing_2019, kim_highly_2019}. However, such approaches sacrifice the ability to adjust footsteps in response to the environment. To combat this rigidity, some methods augment nominal footstep positions through learned networks \cite{villarreal_fast_2019, bratta_contactnet_2023}, nonlinear programs \cite{jenelten_perceptive_2020, grandia_perceptive_2022}, and control barrier functions \cite{grandia_multi-layered_2021}. Through simplifications such as pre-defined gait sequences or discrete search spaces for footholds, these approaches can also be run in real-time.

Some methods resolve all elements of footstep planning (gait sequences, contact positions, and whole-body trajectories) in one module, often through contact-timing optimization \cite{winkler_gait_2018}, mixed-integer programs \cite{deits_footstep_2014, aceituno-cabezas_simultaneous_2018}, soft contact modeling \cite{neunert_whole-body_2018}, or linear complementarity constraints \cite{posa_direct_2014}. While such approaches generate complex, highly dynamic motions, running such planners online is out of the question.

Instead of solving a joint optimization problem, many contact planning frameworks \cite{bretl_motion_2006, hauser_motion_2008, zucker_optimization_2011} employ a hierarchical planning structure, making the key design choice of selecting contacts first and then synthesizing whole-body motions. For bipedal platforms, contact transition models are limited enough to achieve real-time footstep planning through either \textit{pure} search \cite{chestnutt_footstep_2005, griffin_footstep_2019, karkowski_real-time_2016, missura_fast_2021} or \textit{pure} optimization \cite{deits_footstep_2014}. Due to the more aggressive scaling in quadrupedal planning, it becomes crucial to cater particular planning approaches to particular subproblems. 



\subsection{Multi-modal Planning}
A common approach for framing complex long-horizon tasks is through a discrete-continuous or multi-modal motion planning (MMMP) formulation. 
For manipulation, a mode typically corresponds to a particular contact or grasp configuration between end effectors and objects \cite{jorgensen_finding_2020, kim_reachability_2023, chen_trajectotree_2021, kingston_scaling_2023, sleiman_versatile_2023}. On the other hand, in the realm of locomotion, a mode may represent a contact configuration \cite{hauser_motion_2008}, a gait \cite{amatucci_monte_2022, ubellacker_robust_2022}, or a motion primitive \cite{terry_suh_energy-efficient_2020}. In the context of TAMP, these modes are often represented by symbolic states or logic rules depending on particular problem domains \cite{toussaint_differentiable_2018, hartmann_long-horizon_2023, zhao_sydebo_2021, kim_reachability_2023}. 

Existing MMMP frameworks often search over a mode graph which defines valid transitions to obtain mode sequences. Continuous motion planning, whether it be sampling, spline generation, or TO is then performed at the lower level to resolve mode transitions. The works of \cite{toussaint_multi-bound_2017, zhao_sydebo_2021, amatucci_monte_2022} use the results of lower-level TO programs to inform graph edge weights. Other approaches train neural networks to estimate system dynamics \cite{lin_efficient_2019}, the costs of TO programs \cite{terry_suh_energy-efficient_2020}, or the feasibility of a candidate action \cite{driess_deep_2020, driess_deep_2020-1, wells_learning_2019}.

One particular heuristic of interest is the experience-based framework ALEF \cite{kingston_scaling_2023} which exploits the implicit continuous manifolds that arise from contact constraints to disperse the results of offline planning queries throughout the mode graph. This makes for a sample-efficient framework that can apply informed weights to unvisited contact transitions. However, many such heuristics, including that of experience, have yet to be leveraged in quadrupedal contact planning. Recent work \cite{sleiman_versatile_2023} has shown how TAMP and MMMP can enable quadrupeds to rapidly perform complex loco-manipulation tasks such as manipulating and passing through a door. 

\section{Preliminaries} \label{sec:problemformulation}
\subsection{Centroidal Dynamics Model}
A quadrupedal locomotion model can be modeled by centroidal dynamics which bridges the complex full-body dynamics and simple center-of-mass (CoM) dynamics. 
This model constrains the rate of centroidal momentum to be:
\begin{equation}\label{eq:cenDyn}
    \mathbf{\dot{h}} = \begin{bmatrix} \mathbf{\dot{k}} \\ \mathbf{\dot{l}} \end{bmatrix} =
    \begin{bmatrix} \sum_l \mathbf{f}_l + m \mathbf{g} \\ \sum_l (\mathbf{c}_l - \mathbf{r}) \times \mathbf{f}_l \end{bmatrix}
\end{equation}
where $\mathbf{h} = [\mathbf{k}, \mathbf{l}]^T \in \mathbb{R}^6$ is the centroidal momentum which includes linear $\mathbf{k} \in \mathbb{R}^3$ and angular $\mathbf{l} \in \mathbb{R}^3$ momentum, $m$ is the robot mass, $\mathbf{r} \in \mathbb{R}^3$ is the robot CoM position, $\mathbf{f}_l \in \mathbb{R}^3$ is the contact force at the $l^{\rm th}$ foot, $\mathbf{g} \in \mathbb{R}^3$ is the acceleration vector of gravity, and $\mathbf{c}_l \in \mathbb{R}^3$ is the contact position of foot $l$. By using the centroidal momentum matrix (CMM) $\mathbf{A(q)} \in \mathbb{R}^{6 \times (6 + n_l)}$ \cite{orin_centroidal_2013}, $\mathbf{h}$ can also be expressed as: 
\begin{equation}\label{eq:cmmToGeneralizedVel}
    \mathbf{h} = \underbrace{\begin{bmatrix} \mathbf{A}_{b}(\mathbf{q}) & \mathbf{A}_{j}(\mathbf{q}) \end{bmatrix}}_{\mathbf{A}(\mathbf{q})} \begin{bmatrix} \mathbf{\dot{q}}_{b} \\ \mathbf{\dot{q}}_{j} \end{bmatrix}
\end{equation}
where $\mathbf{q} = [\mathbf{q}_b, \mathbf{q}_j]^T \in \mathbb{R}^{6 + n_j}$ is the robot configuration, with $\mathbf{q}_b \in \mathbb{R}^6$ as the floating base pose and $\mathbf{q}_j \in \mathbb{R}^{n_j}$ as the configuration of $n_j$ joints.

\subsection{Contact Manifolds}
In this section, we present the manifold terminologies from the ALEF framework in the context of quadrupedal locomotion. More details can be found at \cite{kingston_scaling_2023}.

In legged locomotion, a contact \textit{mode} $\xi$ can be viewed as a set of footholds at unique positions along a set of steppable objects such as planar stepping stones or linear rebar poles. From this contact mode $\xi$, a lower-dimensional mode manifold $\mathcal{M}^\xi$ embedded in the configuration space $\mathcal{Q}$ arises which encompasses all of the whole-body configurations that satisfy the foothold positions. As the foothold positions vary along the steppable objects, different contact manifolds arise.


The set of all contact modes corresponding to the same set of steppable objects can be grouped into a \textit{mode family} $\Xi$. From each mode familiy $\Xi_i$ arises a foliated manifold, or a foliation $\mathcal{F}_{\Xi_i}$. An $n$-dimensional foliation $\mathcal{M}$ is a manifold defined by a $n_{\boldsymbol{\chi}}$-dimensional \textit{transverse manifold} $X$, a set of non-overlapping $(n-n_
{\boldsymbol{\chi}})$-dimensional $\textit{leaf manifolds}$ $\mathcal{L}_{\boldsymbol{\chi}}$ $\forall \boldsymbol{\chi} \in X$, and lastly a projection operator $\pi: \mathcal{M} \rightarrow X$.


Elements of the transverse manifold $\boldsymbol{\chi} \in X$ are called \textit{coparameters}. A coparameter $\boldsymbol{\chi}$ uniquely parameterizes a mode $\xi$ (and subsequently a leaf manifold $\mathcal{L_{\boldsymbol{\chi}}} = \mathcal{M}^\xi$), and for our use case, the foothold positions along the steppable objects within the given mode family. Therefore, a mode $\xi = \langle \Xi , \boldsymbol{\chi} \rangle$ can be viewed as the tuple of a mode family and a coparameter. The union of leaf manifolds along the set of coparameters $\bigcup_{\boldsymbol{\chi} \in X}  \mathcal{L}_{\boldsymbol{\chi}}$ recovers the entire foliation $\mathcal{M}$.

A leaf or mode manifold $\mathcal{M}^\xi$ can be implicitly defined through a constraint function $F^\xi : \mathbb{R}^n \rightarrow \mathbb{R}^{n_{\boldsymbol{\chi}}}$ where a configuration $\mathbf{q}$ lies on the mode manifold if $F^\xi(\mathbf{q}) = \mathbf{0}$.

In the proposed work, a mode represents three stance legs in contact with three separate rebars. Therefore, an example constraint function for a contact mode is
\begin{equation}
    F^\xi(\mathbf{q}) := \begin{bmatrix}
                    F^\xi_{1}(\mathbf{q}), 
                    F^\xi_{2}(\mathbf{q}), 
                    F^\xi_{3}(\mathbf{q})
                \end{bmatrix}^T,
\end{equation}
where for a foot $l$ in contact with a bar defined by a starting point $\mathbf{p}_0 \in \mathbb{R}^3$ and ending point $\mathbf{p}_1 \in \mathbb{R}^3$,
\begin{equation}
     F^{\xi}_{l}(\mathbf{q}) := 
     \text{FK}_{l}(\mathbf{q}) - (\mathbf{p}_0 + \boldsymbol{\chi}_{l}(\mathbf{p}_1 - \mathbf{p}_0))
    = \textbf{0},
\end{equation}
where $\text{FK}_{l}(\mathbf{q})$ gives the position of foot $l$ given $\mathbf{q}$ via forward kinematics (FK). In this work, we assume that the rebar grid is planar and aligned with the ground plane, and we assume that the poses of the rebars are known.
\section{Multi-modal Planning} \label{sec:mmp}
Assume a quadrupedal robot with a configuration space $\mathcal{Q} \subset \mathbb{R}^{6 + n_j}$. We seek to find a collision-free path $\mathbf{q}(s)$ with $s \in [0, 1]$ from $\mathbf{q}(0) = \mathbf{q}_{\rm start}$ to $\mathbf{q}(1) = \mathbf{q}_{\rm goal}$ in which contact must strictly be made with the rebar grid. 

\subsection{Mode Transition Graph Construction} \label{sec:graph_construction}
We employ a mode transition graph $\mathcal{G} = (\mathcal{V}, \mathcal{E})$ as in the ALEF framework in which the mode families comprise the set of vertices $\mathcal{V}$. The edges $\mathcal{E}$ are then formed between mode families for which kinematically feasible transitions exist. In the configuration space, edges are formed between mode families which have foliations that intersect.

\begin{figure}
    \centering
    \includegraphics[width=0.35\textwidth]{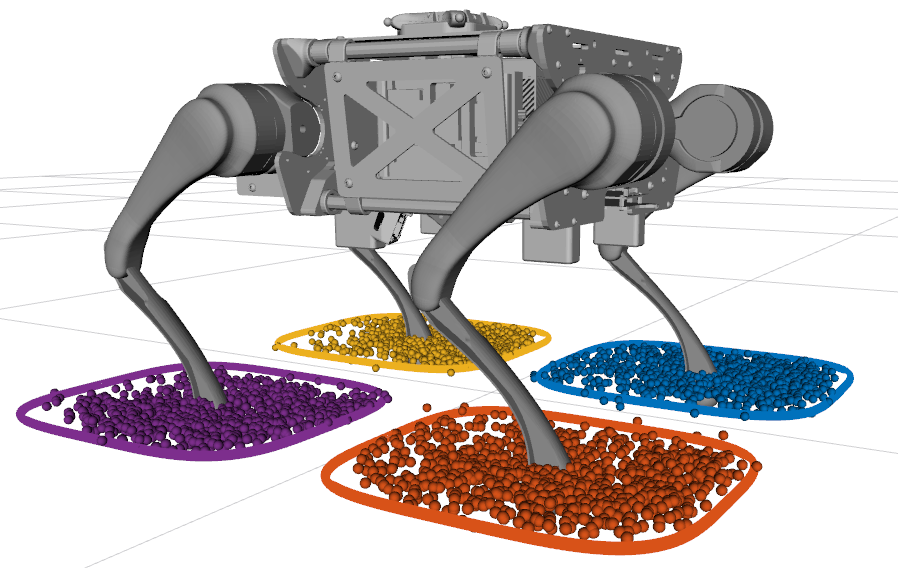}
    \caption{Kinematic reachability areas along with projected configuration samples. Some samples that meet the distance threshold are set as outliers to avoid reachable areas that yield unstable contacts or self-collisions.}
    \label{fig:superquadrics_fig}
    \vspace{-0.25in}
\end{figure}

\begin{figure}[t]
    \vspace{0.2cm}
    \centering
    \includegraphics[width=0.48\textwidth]{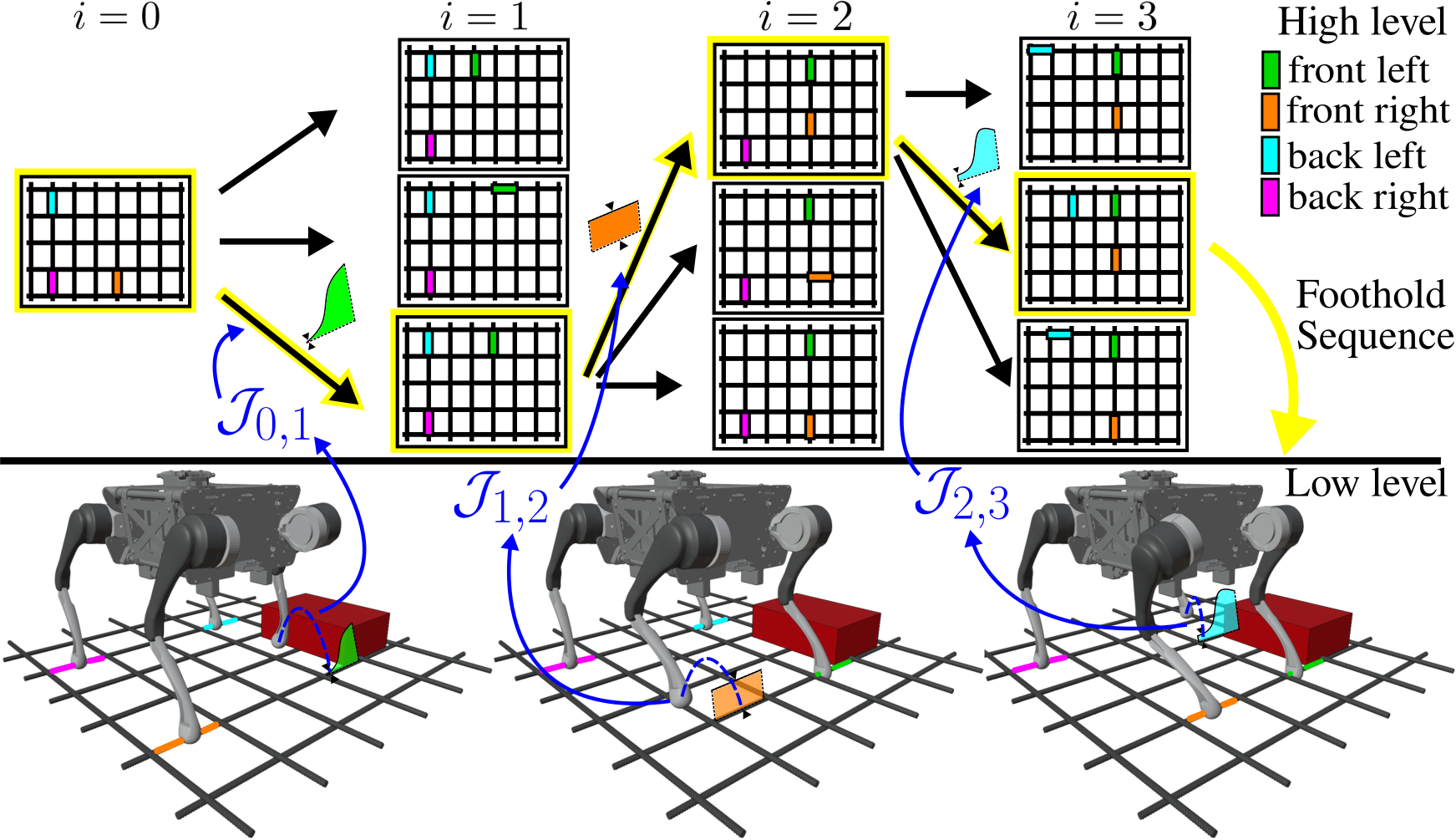}
    \caption{Overall diagram of planning framework. A graph search (Section \ref{sec:graph_search} is performed over mode transitions using the estimated weight distributions from experience (Section \ref{sec:experience}) which are colored by their corresponding swing foot. Then, the suggested contact sequence is run through trajectory optimization (Section \ref{sec:traj_opt}) to determine the costs $\mathcal{J}_{i, i+1}$ of the transitions.}
    \label{fig:diagram}
    \vspace{-0.5cm}
\end{figure}

We utilize locomotion-specific constraints to implicitly define feasible transitions within the mode transition graph. First, a user-defined contact sequence informs the graph on the robot's footfall pattern. A transition from a vertex $v_i$ to vertex $v_{i+1}$ (equivalently, mode family $\Xi_i$ to mode family $\Xi_{i+1}$) is only added if the swing leg at $v_i$ and the swing leg at $v_{i+1}$ occur sequentially in the contact sequence. Second, we incorporate kinematic reachability analysis to approximate what regions are contactable by the robot's legs. 
In this work, we use a family of functions known as superquadrics that has seen recent use in legged locomotion \cite{melon_receding-horizon_2021}. The set of points that fall within the superquadric centered at $(x_0, y_0)$ are
\begin{equation}
    S = \big\{ (x,y) \in \mathbb{R}^2 \hspace{0.25cm} \Big| \hspace{0.25cm} \big| \frac{x - x_0}{A} \big| ^a + \big| \frac{y - y_0}{B} \big| ^b \leq 1 \big\},
\end{equation}
where scalars $A, B$ and $a, b$ control dimensions and curvature respectively. Parameters were obtained through randomly sampling configurations, keeping all samples that reach within a distance threshold $\epsilon = 1$ cm of the ground, and tuning parameter values to encompass the samples in contact (see Figure \ref{fig:superquadrics_fig}). For clarity, a portion of the mode transition graph with these constraints incorporated into the feasible transitions is visualized within Figure \ref{fig:diagram}.

\subsection{Mode Transition Graph Search} \label{sec:graph_search}
We formulate the task of finding a discrete foothold sequence as a graph search problem over the aforementioned mode transition graph. We employ the A* search algorithm for this mode transition graph search. For the search, we discretize along the \textit{transverse manifold} of each mode family to generate \say{slices} of the foliations that correspond to intervals of foothold positioning. Edges are then added between all slices of the source and destination mode families in which contact transitions exist in the graph. This discretization allows for the search to reason about both contact sequencing (which rebars to contact with which feet) and foothold sequencing (where to make contact). This search provides a candidate \textit{lead} --- a sequence of modes $[ \xi_0, \xi_1, \dotsc,  \xi_i, \xi_{i+1}, \dotsc, \xi_M]$ where $M$ is the length of the lead --- which defines a foothold sequence from start to goal. 

\subsubsection{Edge weight} For a transition between source mode $\xi_{i} = \langle \Xi_{i}, \boldsymbol{\chi}_{i} \rangle$ and destination mode $\xi_{i+1} = \langle \Xi_{i+1}, \boldsymbol{\chi}_{i+1} \rangle$, the graph edge $e = (\xi_{i}, \xi_{i+1})$ is assigned the weight
\begin{equation}
\begin{split}
& \Delta c(\xi_{i}, \xi_{i+1}) = w_\mathcal{D} \cdot \mathcal{D}^{\Xi_{i},\Xi_{i+1}}(\boldsymbol{\chi}_{i}, \boldsymbol{\chi}_{i+1}) + \\
& w_d \cdot d_{\text{CoM}}(\xi_{i}, \xi_{i+1}) + w_\tau \cdot d_{\tau}(\xi_{i}, \xi_{i+1}),
\end{split}
\end{equation}
where the distribution $\mathcal{D}^{\Xi_{i},\Xi_{i+1}}(\chi_{i}, \chi_{i+1})$ captures the difficulty of transitioning from $\xi_{i}$ to $\xi_{i+1}$. This distribution is estimated offline through the experience heuristic which is detailed in Section \ref{sec:experience}. The term $d_{\text{CoM}}(\xi_{i}, \xi_{i+1})$ is the Euclidean distance between nominal CoM positions for $\xi_{i}$ and $\xi_{i+1}$, and $d_{\tau}(\xi_{i}, \xi_{i+1})$ is the deviation of nominal CoM positions for $\xi_{i}$ and $\xi_{i+1}$ from a suggested torso path. This path can come from any planner that generates a sequence of torso poses from start to goal, and implementation details on this suggested torso path are given in Section \ref{sec:experimentalresults}.


\subsubsection{Cost-to-go} For the A* search, a contact mode $\xi = \langle \Xi, \chi \rangle$ is assigned the search heuristic value  
\begin{equation}
    g(\xi) = w_d \cdot d_{\text{CoM}}(\xi, \Xi_{\rm goal})
\end{equation}
to ensure an admissible search heuristic and therefore provide optimal foothold sequences with respect to edge weights.

\subsection{Whole-Body Trajectory Optimization} \label{sec:traj_opt}
Once a discrete lead is obtained, continuous motions plans must be synthesized between the consecutive modes. For this, we opt to use trajectory optimization (TO) as opposed to the sampling-based planning methods \cite{simeon_manipulation_2004, gravot_asymov_2005, mirabel_manipulation_2017} commonly used in MMMP in order to be able to generate continuous dynamics-aware paths that enable the robot to transition between modes. While the TO sacrifices the probabilistic completeness that is offered by sampling-based methods, our approach enables kinodynamically-aware multi-modal contact planning which has not been explored in prior works. Similar to \cite{dai_whole-body_2014,chignoli_mit_2021,sleiman_unified_2021}, our TO problem solves over the robot state $\mathbf{x} = [\mathbf{h}, \mathbf{q}_b, \mathbf{q}_j]$, 
the robot input $\mathbf{u} = [\mathbf{f}, \mathbf{v}_j]$,
and the coparameters $\boldsymbol{\chi}_{i+1}$ of the destination mode family. 
The TO formulation for generating a whole-body trajectory to transition from $\xi_{i} = \langle \Xi_{i}, \boldsymbol{\chi}_{i} \rangle$ to $\xi_{i+1} = \langle \Xi_{i+1}, \boldsymbol{\chi}_{i+1} \rangle$ is written as:
\begin{subequations} \label{eq:whole_to}
\begin{align} 
& &&  \| \mathbf{x}[N] - \mathbf{x}^{\text{des}}[N] \|^2 _{Q_f} +  &&& \nonumber \\
& \underset{\mathbf{x},\mathbf{u}, \boldsymbol{\chi}_{i+1}}{\min} && \sum_{k=0}^{N - 1} \Biggl( \| \mathbf{x}[k] - \mathbf{x}^{\text{des}}[k] \|^2 _Q + \| \mathbf{u}[k] \|^2_{R} \Biggr) &&& \nonumber \\
& \text{subject to}  && &&& \nonumber \\
& \text{(Mode $i$)} && F^{\xi_{i}}(\mathbf{q}[k]) = \mathbf{0} &&& \label{eq:src_mode} \\
& \text{(Mode $i+1$)} && F^{\xi_{i+1}}(\mathbf{q}[N], \boldsymbol{\chi}_{i+1}) = \mathbf{0} &&& \label{eq:dst_mode} \\
& \text{(Dynamics)} && \begin{bmatrix} \mathbf{\dot{l}}[k] \\ \mathbf{\dot{k}}[k] \\ \mathbf{\dot{q}}_b[k] \\ \mathbf{\dot{q}}_j[k]\end{bmatrix} = 
    \begin{bmatrix} \sum_l \mathbf{f}_l[k] + m \mathbf{g} \\ 
    \sum_l (\mathbf{c}_l - \mathbf{r}) \times \mathbf{f}_l[k] \vspace{0.1cm}\\
    \mathbf{A}_b^{-1}(\mathbf{h}[k] - \mathbf{A}_j \mathbf{v}_j[k]) \\ \mathbf{v}_j[k] \end{bmatrix} \label{eq:dynamics} \\
& \text{(Friction)} && \mathbf{f}_l[k] \in \mathcal{F}_l(\mu, \mathbf{q}) \hspace{1cm} \forall l \in \mathcal{C}_{i} &&& \label{eq:friction_cone} \\
& && \mathbf{f}_l[k] = \mathbf{0} \hspace{1cm} \quad \quad \quad \forall l \notin \mathcal{C}_{i} &&& \label{eq:zero_force} \\
& \text{(Collision)} && g(\mathbf{q}[k]) \geq 0  \quad \quad \quad \; \forall \hspace{0.05cm} k \in [0, N] &&& \label{eq:collision_avoidance} 
\end{align}
\end{subequations}
where $\mathcal{F}_l(\mu, \mathbf{q})$ represents the friction cone which depends on the friction coefficient $\mu$ and robot pose $\mathbf{q}$, and $\mathcal{C}_{i}$ represents the set of stance feet for $\xi_{i}$. Note that at the TO level, the destination mode family is fixed, but the coparameter is treated as a decision variable, allowing for variation of the destination mode.
We formulate the above TO as a Sequential Quadratic Program (SQP) and solve through
the OCS2 library \cite{farbod_farshidian_and_others_ocs2_nodate}. We employ a time horizon of $T = 0.5$ seconds and $N=50$ knot points with maximal 250 iterations.
For the collision avoidance constraint, we run the Gilbert-Johnson-Keerthi algorithm \cite{gilbert_fast_1988} provided by the HPP-FCL library \cite{noauthor_hpp-fcl_nodate}. We only check for collisions with the robot's feet to reduce computation time. We also use the Pinocchio library for kinematics and dynamics calculations \cite{carpentier_pinocchio_2019}. $\mathbf{x}^{\rm des}$ is generated in two steps, first, the mode constraint functions from constraints (\ref{eq:src_mode}) and (\ref{eq:dst_mode}) are used in a contact projection step to project a randomly sampled target configuration into contact satisfying the source and destination modes. If this first step is successful, cubic splines are then synthesized for the swing feet to build out the remainder of $\mathbf{x}^{\rm des}$. If the optimal cost of an attempted mode transition is above a threshold $\mathcal{J}_{\rm max}$, then the planning trial is terminated.

\section{Planning with Experience} \label{sec:experience}

The objective of planning with experience is to acquire a continuous function that captures the difficulty or cost of attempting certain contact transitions within the environment which are encoded as the edges of our mode transition graph.


\subsection{Optimal Cost Integration}
There are two potential outcomes of a transition attempt:
\begin{enumerate}
    \item Contact projection fails to generate a target configuration, suggesting an infeasible transition ($\mathcal{J}_{i, i+1} = \infty$)
    \item Contact projection generates a target configuration, triggering a TO instance ($\mathcal{J}_{i, i+1} = \text{Equation \ref{eq:whole_to}}$)
\end{enumerate}

The experience heuristic allows us to infer from previously attempted mode transitions the cost of nearby, possibly unattempted mode transitions. While a transition between two modes may be kinematically infeasible, infeasibility does not necessarily hold for all modes between the two mode families. Therefore, weighting the entire transition with a cost of infinity could inhibit discovery of a path to the goal, especially in situations where foothold location is crucial to successful locomotion. 
To account for this, cost values $\mathcal{J}_{i, i+1}$ are first passed through a weighted tanh function
\begin{equation}
    \delta_{i, i+1} = w_1 \tanh{(w_2 \cdot \mathcal{J}_{i, i+1} + w_3)},
\end{equation}
where $w_1$, $w_2$, and $w_3$ are positive scalar values to map the costs to finite positive \textit{penalties} that can be used to populate the edge weights in the mode transition graph. 
\subsection{Experience Accumulation}
The smoothness of contact manifolds allows us to exploit prior planning results to inform estimates regarding similar contact transitions. This is predicated on the idea that since foliations are smooth, coparameters nearby on the transverse manifold parameterize modes that are similar in cost \cite{kingston_scaling_2023}.

Once the penalty values are obtained from a given TO run, they can be distributed throughout the graph in the form of experience. To distribute penalties throughout all modes within a given mode family, we employ function regression techniques that involve constructing a weighted sum of basis functions that is meant to estimate the continuous distribution of the average penalty value at different contact positions. In this work, we model this distribution as the weighted sum of radial basis functions (RBFs), where the update applied to the weights of the traversed edge is
\begin{equation}
    f^{\Xi_{i}, \Xi_{i+1}}(\boldsymbol{\chi}_{i}, \boldsymbol{\chi}_{i+1}) = w_e \cdot \exp(\frac{-d(\boldsymbol{\chi}_{i}, \boldsymbol{\chi}_{i+1})^2}{2 \cdot \sigma^2})
\end{equation}
where
\begin{equation}
    w_e = (\mathcal{J}_{i, i+1} - \Bar{\mathcal{J}})
\end{equation}
where $\mathcal{J}_{i, i+1}$ is the cost obtained from the attempted mode transition, $\Bar{\mathcal{J}}$ is the average transition cost between $\xi_{i}$ and $\xi_{i+1}$, $d(\boldsymbol{\chi}_{i}, \boldsymbol{\chi}_{i+1})$ represents the distance of a coparameter to $(\boldsymbol{\chi}_{i}, \boldsymbol{\chi}_{i+1})$, and $\sigma$ represents the standard deviation of the RBF. This update adds a basis function to the weight distribution of each traversed edge that is centered at the attempted coparameter value and weighted by its deviation from the average cost of the attempted mode transition. 

\section{Experimental Results} \label{sec:experimentalresults}


In this section, we perform offline experience accumulation in which a batch of planning trials are run in order to populate the graph edge weights and demonstrate the whole-body motion plans output by the proposed framework. Within each planning trial, the high-level graph search provides a candidate contact sequence which is passed to the lower level of the framework where a sequence of trajectory optimization subproblems are solved to generate contact transitions. 

\begin{figure}[h!]
    \centering
        \vspace{-0.3cm}
\includegraphics[width=0.48\textwidth]{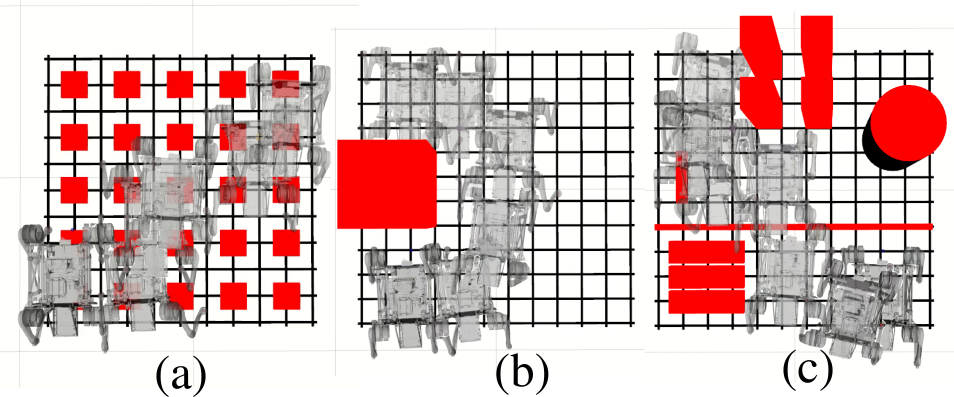}
    \caption{Rebar grid layouts used in the case studies in Sections \ref{sec:cs1_exp} - \ref{sec:cs3_construction}.}
    \label{fig:exp2_setups}
    \vspace{-0.3cm}
\end{figure}

Case studies in three environments are performed: (i) a grid with low-height obstacles scattered along its surface (Section \ref{sec:cs1_exp}), (ii) a grid with a tall obstacle positioned between the start and goal configurations (Section \ref{sec:cs2_guiding_torso_path}), and (iii) a grid with various obstacles meant to emulate a real-world constriction site. These three grids, along with outputted reference trajectories, are visualized  in Figure \ref{fig:exp2_setups}. 

For the three case studies, we record computation times of the graph search triggered for each planning trial as well as average, minimum, and maximum TO solve times across all of the attempted subproblems within each planning trial. Additionally, we report the results of all subproblems -- success, failure, or not attempted due to early trial termination -- as well as the total path costs for the trials that reached the goal. Lastly, reference trajectories obtained from our framework are deployed on a quadruped on a real world rebar grid and tracking performance is evaluated (Section \ref{sec:cs4_ref_traj_tracking}). 

\subsection{Footstep Adjustment through Experience}
\label{sec:cs1_exp}
\begin{figure}[h]
    \vspace{-0.2cm}
    \centering
    \includegraphics[width=0.48\textwidth]{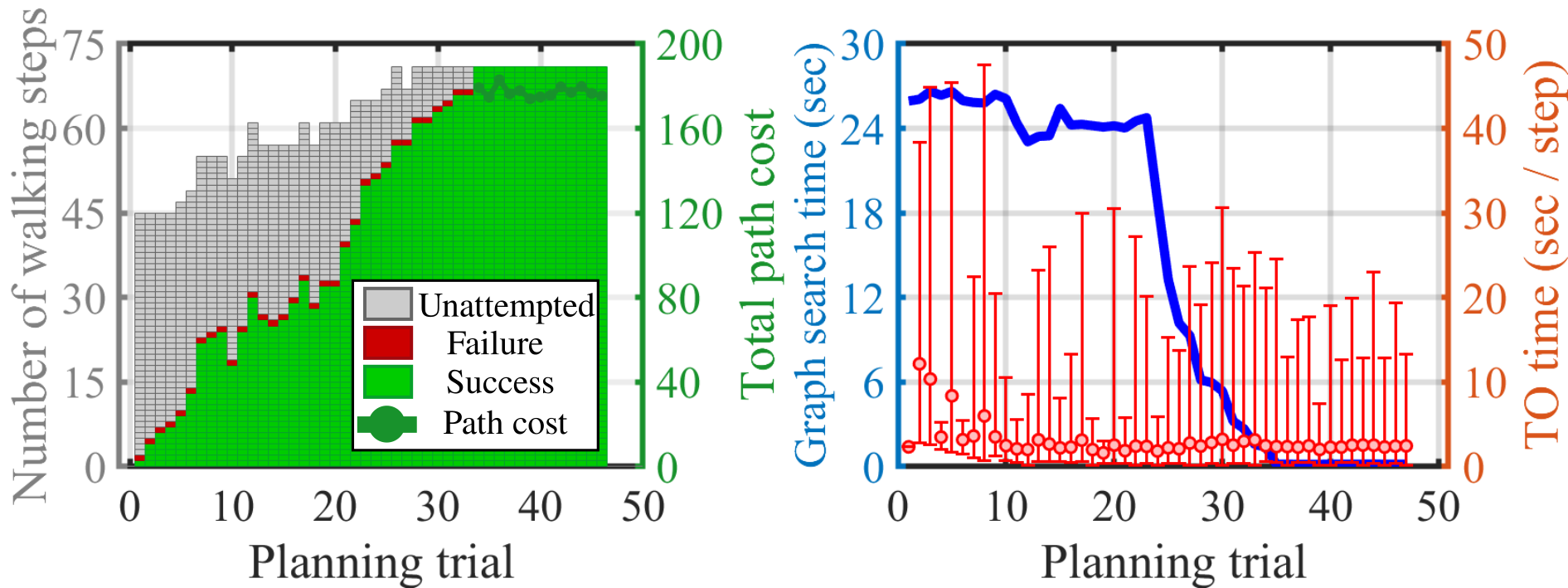}
    \caption{Results for Section \ref{sec:cs1_exp}: Case Study 1 on the grid in Figure \ref{fig:exp2_setups}(a). A gradual increase in progress through the proposed contact sequences can be seen in left figure as the graph search comes to avoid the more difficult contact transitions that lead to collisions. Once successful contact sequences are identified through experience, the graph search time drops significantly.}
    \label{fig:exp2_plots}
    \vspace{-0.7cm}
\end{figure}

In this first case study, we demonstrate the key role that the weight distributions obtained through experience play in successful contact planning. We deploy the planner on a rebar grid with short, foot-level obstacles along its surface, and through offline experience accumulation the planner ascertains what contact transitions allow the robot to reach the goal without collisions. Results are shown in Figure \ref{fig:exp2_plots}.

We initially performed 500 planning trials with all experience-based weight distributions initialized uniformly to $\mathcal{D}^{\Xi_{i},\Xi_{i+1}}(\boldsymbol{\chi}_{i}, \boldsymbol{\chi}_{i+1}) = 0.01$, but the planner failed to reach the goal on any of the trials due to the extensive period of graph exploration required to appropriately estimate the weight distributions. We then performed a second run of offline trials where we initialized the weight distributions to priors based on proximity of the mode transitions to obstacles in the environment. With these priors, the planner was able to explore a greater portion of the mode transition graph and ultimately find successful contact plans to the goal in far fewer trials. During initial trials, mode transitions that collide with obstacles are attempted, leading to extremely prohibitive and highly variant TO times. However, after roughly $30$ trials the planner is able to suggest collision-free contact plans, greatly reducing both the mean and variance of TO solve times. In this environment, the torso planner does not provide any useful insights on planning given that all obstacles exist at the foot level, and the key heuristic that enables successful planning to the goal in such an environment is the weight distribution accumulated from experience.

\subsection{Torso Path-Guided Experience Accumulation}
\label{sec:cs2_guiding_torso_path}
\begin{figure}[h]
    \centering
    \begin{subfigure}[b]{0.45\textwidth}
        \includegraphics[width=\textwidth]{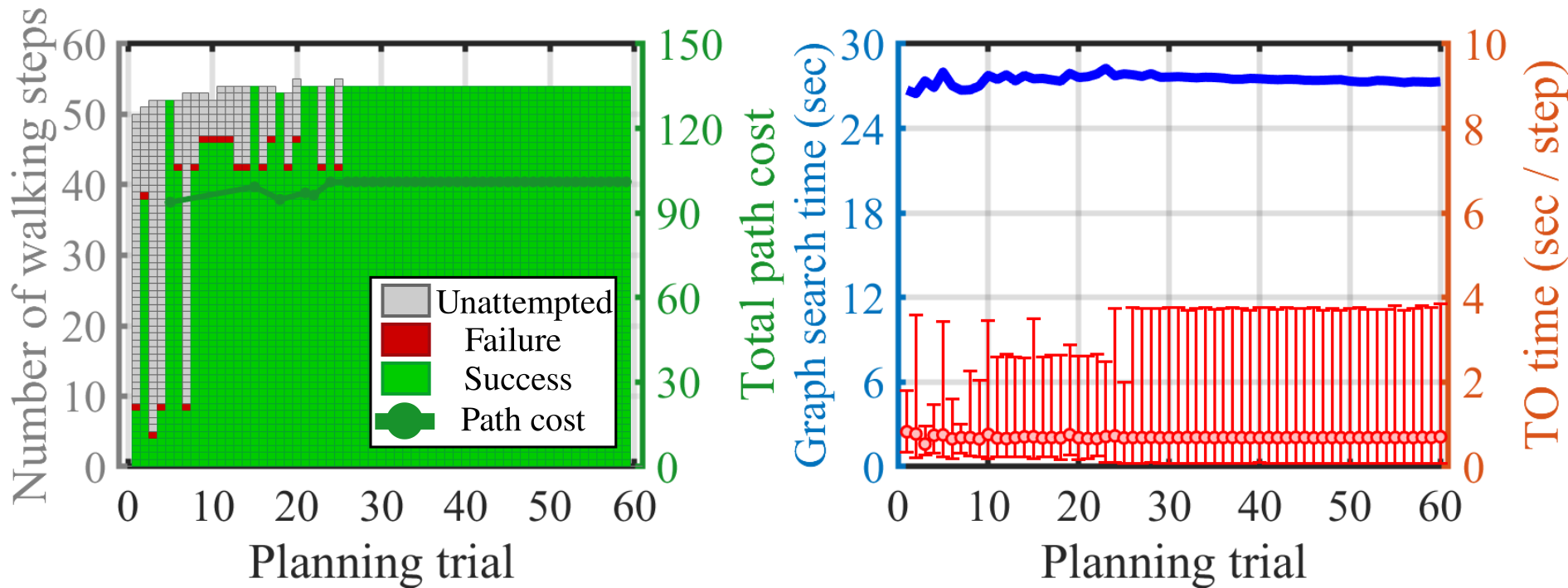}
        \caption{Results for experience accumulation with the guiding torso path.}
        \label{fig:exp1_torso_plots}
    \end{subfigure}
    \hfill
    \begin{subfigure}[b]{0.49\textwidth}
        \includegraphics[width=\textwidth]{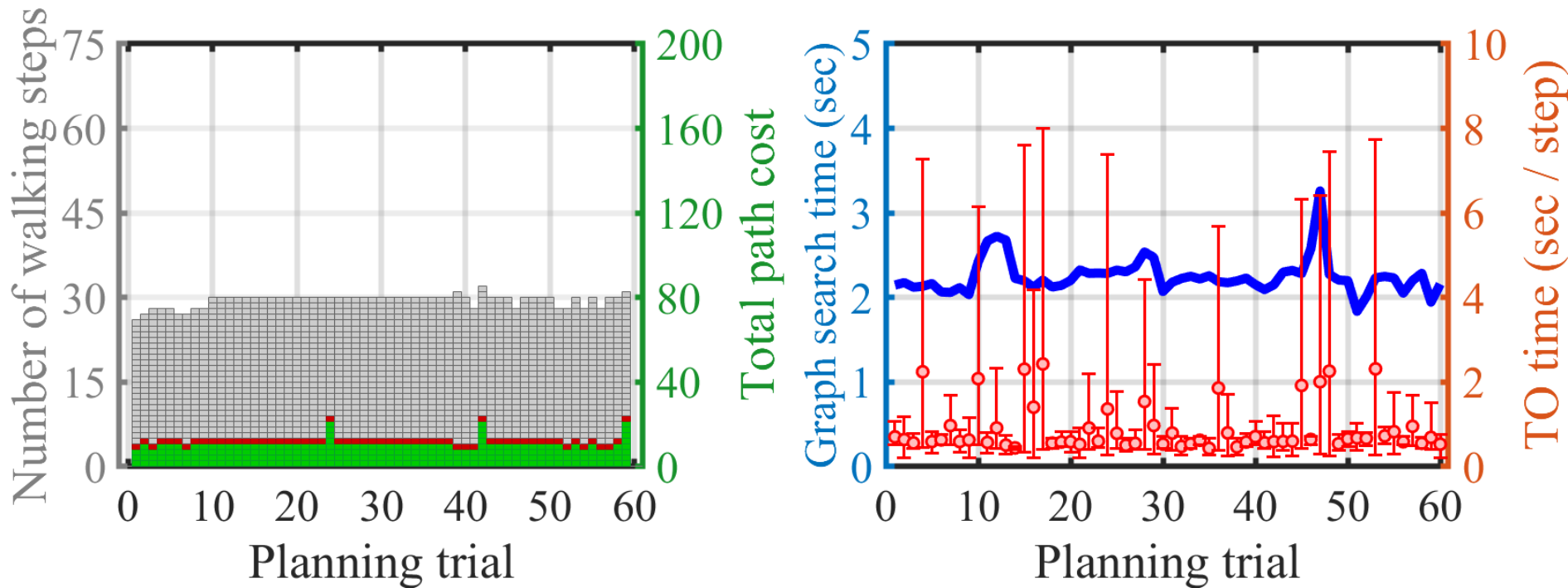}
        \caption{Results for experience accumulation without the guiding torso path.}
        \label{fig:exp1_no_torso_plots}
    \end{subfigure}
    \caption{Results for Section \ref{sec:cs2_guiding_torso_path}: Case Study 2 on the grid in Figure \ref{fig:exp2_setups}(b). (a) The torso path directs the search away from obstacles and towards the richer areas of the environment for footholds, leading to fewer required offline trials. (b) Without the guiding torso path, the contact planner initially searches for footholds along a straight path between the start and the goal which leads to a significantly longer experience accumulation process. }
    \label{fig:exp1_results}
    \vspace{-0.6cm}
\end{figure}

In this section, we perform an ablation study in which the multi-modal contact planner is run both with and without a guiding torso path planner. The incorporation of this planner emulates common navigation frameworks which perform coarse, low-frequency torso planning that provides guiding paths to a lower-level footstep planner that synthesizes whole-body trajectories. We formulate the torso path planner as an additional A* graph search where the graph nodes are a set of positions along the rebar grid. Edges are added between nodes that are under a distance threshold $\tau = 0.15$ m apart, and edge  weights are assigned based on proximity to obstacles. Results are shown in Figures \ref{fig:exp1_torso_plots} and \ref{fig:exp1_no_torso_plots}. 

Due to the large obstacle positioned between the start and goal, the added torso planner greatly expedites experience accumulation. The guiding torso path biases the multi-modal contact planner towards obstacle-free regions of the grid which take far less time to plan through with TO. Without the torso path, the mode transition graph search takes the shortest path from start to goal which runs through the obstacle, leading to much higher TO times and less overall exploration of the graph. One drawback of instituting the guiding torso path is that the graph search times increase significantly due to the introduction of the complicated torso path deviation term into the graph edge weight function.

\subsection{Holistic Collision Avoidance through Experience}
\label{sec:cs3_construction}

\begin{figure}[h]
    \centering
        \vspace{-0.2cm}
    \includegraphics[width=0.48\textwidth]{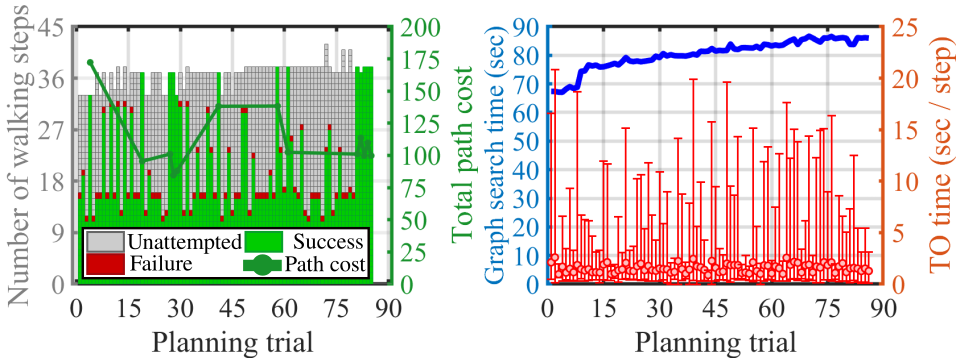}
    \caption{Results for Section \ref{sec:cs3_construction}: Case Study 3 on the grid in Figure \ref{fig:exp2_setups}(c). With this more complicated environment came a longer experience accumulation process as well as longer graph search times due to the different weighting terms of the graph search interacting in nontrivial ways.}
    \label{fig:exp3_plots}
    \vspace{-0.4cm}
\end{figure}

In the third case study, we demonstrate our framework's ability to reason through complicated rebar environments with obstacles at both the torso level and the foot level. Larger pillars and beams are avoided through following the guiding torso path while barriers and debris on the grid surface are avoided through adjusting footstep positions through experience. Results are shown in Figure \ref{fig:exp3_plots}.

In this study, there are some early planning trials that successfully reach the goal. This is largely due to the presence of the guiding torso path manuevering the resulting contact sequences around large obstacles. However, over the course of the experience accumulation, the mode transitions that allow the robot to step over the barrier spanning across the grid and avoid the clutter on the left side of the grid are discovered and exploited. This complicated environment gives rise to higher graph search times than those observed in the previous case studies. Also, more planning trials are required to appropriately estimate the edge weights within the graph. However, our framework still only requires $80$ trials to generate contact sequences with consistently short solve times at the TO level that allow robot to reach the goal.

\subsection{Hardware Implementation}
\label{sec:cs4_ref_traj_tracking}
To ensure that trajectories generated by our framework can be robustly deployed on real systems, we set up a rebar scenario and performed trials onboard a quadrupedal rebar-tying robot -- Chotu. We employed an MPC-WBC tracking controller modified from \cite{sleiman_unified_2021}. The MPC solves a similar centroidal dynamics optimization as (\ref{eq:whole_to}) at 100 Hz but with the fixed contact sequence from our framework. An end-effector constraint is added to accurately track the reference swing foot trajectory, which is crucial to successful rebar traversing. The WBC solves a hierarchical QP at 500 Hz. The state estimator fuses IMU data, joint encoders, and motion capture inputs to provide accurate body position information.

\begin{figure}[h!]
    \centering
    \includegraphics[width=\linewidth]{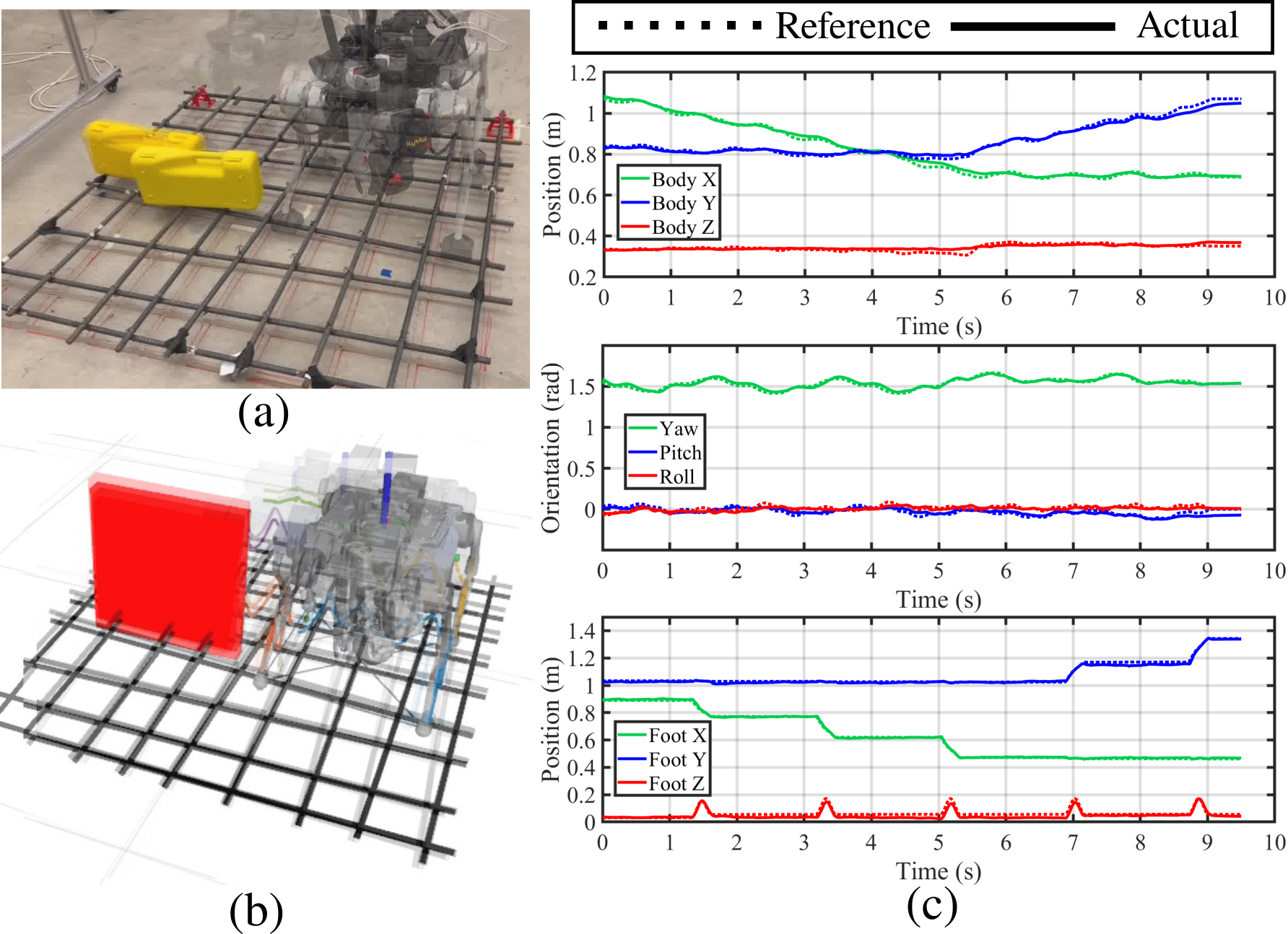}
    \caption{Hardware demonstration of Chotu performing rebar traversal. (a) Real-world rebar grid setup. (b) Rebar traversal in dynamics simulation. Colored lines denote desired position trajectories for robot torso and feet. (c) Comparison of reference trajectory and measured robot states.}
    \label{fig:exp3_spaced_grid_and_plans}
    \vspace{-0.3cm}
\end{figure}

We validate the trajectories generated by our framework on one real-world example. As shown in Fig.~\ref{fig:exp3_spaced_grid_and_plans} (a) and (b), the robot Chotu is commanded to move from the top left corner of the rebar grid to the middle right with an obstacle is blocking in the way Fig.~\ref{fig:exp3_spaced_grid_and_plans} (c) demonstrates a favorable tracking performance with insignificant body pose and foot error regarding the body pose and foot locations.

\section{Conclusion}
\label{sec:Conclusion}

In this work, we adapt an efficient multi-modal contact planner to the task of quadrupedal rebar traversal. We accumulate offline experience to estimate optimal cost distributions using the results of lower-level trajectory optimization instances. This framework assumes that poses of environment objects are known which prohibits integration into an online navigation framework. In the future, we aim to incorporate perception into the graph construction and search to allow for deployment in unknown environments. The experience that is accumulated from trajectory optimization is also tied to the environment in which it is obtained. We seek to re-structure this experience heuristic to allow for experience to be used across environments of varying formats and complexities.

\bibliographystyle{IEEEtran}
\bibliography{references}

\end{document}